\documentclass[11pt]{article}

\usepackage[preprint]{acl}

\usepackage{times}
\usepackage{latexsym}

\usepackage[T1]{fontenc}

\usepackage[utf8]{inputenc}

\usepackage{microtype}

\usepackage{inconsolata}

\usepackage{graphicx}
\usepackage{subcaption}
\usepackage{booktabs}
\usepackage{pifont}
\usepackage{enumitem}
\usepackage{amsmath}
\usepackage{listings}
\lstdefinestyle{compactionprompt}{
    basicstyle=\footnotesize\ttfamily,
    columns=fullflexible,
    keepspaces=true,
    breaklines=true,
    breakautoindent=false,
    breakindent=0pt,
    breakatwhitespace=false,
    showstringspaces=false,
    frame=single,
    rulecolor=\color{black!30},
    aboveskip=0.5em,
    belowskip=0.5em
}

\newcommand{\methodname}{\textsc{MMPact}}

\title{MMPact: Autonomous Multimodal Context Compaction for LLM Agents}

\author{
  \textbf{Feng Luo\textsuperscript{1},}
  \textbf{Yang Zhang\textsuperscript{2},}
  \textbf{Yujian Liu\textsuperscript{1},}
  \textbf{Shiyu Chang\textsuperscript{1},}
\\
  \textbf{Mayank Darbari\textsuperscript{3},}
  \textbf{Liangjie Hong\textsuperscript{3},}
  \textbf{Liang Wu\textsuperscript{3}}
\\
\\
  \textsuperscript{1}UC Santa Barbara, \quad
  \textsuperscript{2}MIT-IBM Watson AI Lab, \quad
  \textsuperscript{3}Nokia
}

\begin{document}
\maketitle
\begin{abstract}
LLM agents rely on context compaction to condense accumulating interaction trajectories into concise summaries for continued task execution within context limits.
For multimodal agents, however, text-only summaries may discard visual details that are difficult to describe or whose relevance to follow-up tasks is not yet known. 
Retaining images with text can preserve these details, but requires deciding which images to retain and how to compress them within a limited token budget.
In this paper, we propose \methodname{}, a Multi-Modal comPaction method that enables an LLM to autonomously decide what information to retain and how to represent it under a given token budget.
To systematically evaluate the effectiveness of multimodal compaction, we further introduce TraVQA, a trajectory-based visual QA benchmark that measures information retention for current and follow-up tasks, with all evaluation questions hidden during compaction.
Experiments show that \methodname{} adapts its modality choices and image compression to the budget, favoring text-only summaries under tight budgets and retaining more images with less downscaling under larger budgets. 
At the comprehensive budget level, \methodname{} improves follow-up QA accuracy over text-only summaries by up to 7.69\%  with Qwen3.8-Flash-Next and 5.53\% with Qwen3.8-27B, while reducing multimodal summary tokens.
Our code is available at \url{https://github.com/UCSB-NLP-Chang/MMPact}.
\end{abstract}

\section{Introduction}
\label{sec:intro}
\begin{figure*}[t]
    \centering
    \includegraphics[
    width=\textwidth,
    trim={0mm 5mm 0mm 5mm},
    clip
]{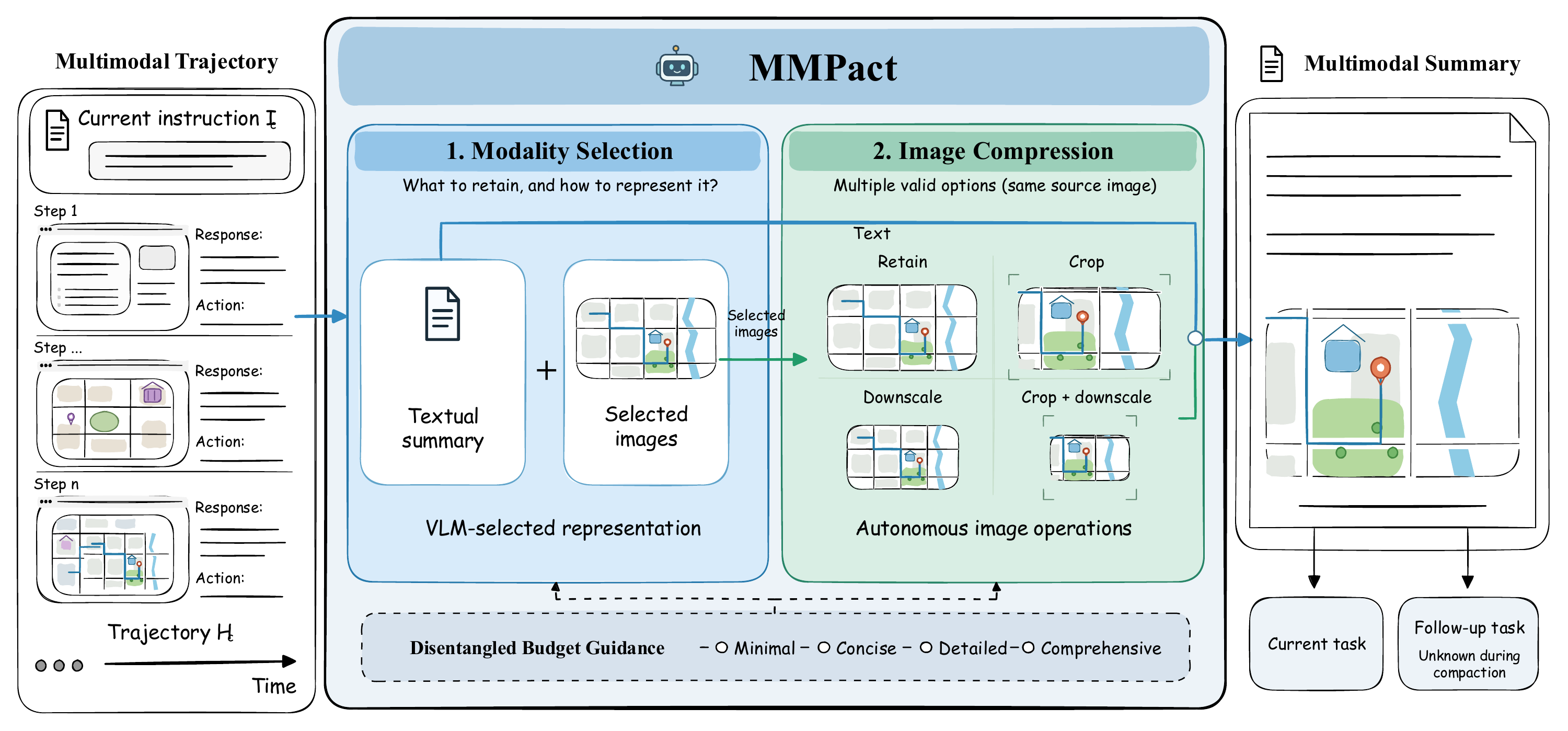}
    \caption{
    Overview of MMPact.
    Given the current instruction $I_c$ and multimodal
    trajectory $H_c$, MMPact produces a multi-modal summary $S$.
    Disentangled budget guidance $b$ specifies the overall
    text-and-image token budget without prescribing a modality
    or compression operation.
    The VLM autonomously selects textual and visual information
    to retain and can retain, crop, downscale, or jointly crop
    and downscale selected images.
    The resulting summary supports the current task and
    follow-up tasks that are unknown during compaction.
    }
    \label{fig:mmpact}
    \vspace{-4mm}
\end{figure*}

Context compaction has become an indispensable component of large language model (LLM) agent harnesses \citep{wang2026openhands,chang2026karl,sengupta2026harbor}. During task execution, LLM agents often accumulate long trajectories consisting of code, tool interactions, and reasoning traces, which can easily exceed the LLM's context window \citep{packer2023memgpt,lu2025scaling,cim2026parallel,zhou2026mem1,min2026adapting}. Compaction addresses this problem by replacing the accumulated trajectory with a condensed summary that preserves the information needed to continue task execution.

For unimodal LLM agents that primarily operate on text and code trajectories, a simple yet effective compaction strategy is to use an LLM to \textbf{summarize} the trajectory into concise text \citep{wu2025resum, li2026compactionrl}. Leveraging the strong text-processing capabilities of LLMs, this strategy offers two advantages: it is \textit{information-sufficient}, preserving the information needed to continue the current task while substantially reducing token usage, and \textit{budget-aware}, allowing the summary length to be adjusted to different token budgets.

However, for multimodal LLM agents \citep{qin2025ui,agashe2025agent,yang2025embodiedbench}, whose trajectories also contain visual observations such as screenshots of graphical interfaces, compaction becomes considerably more challenging, as images contain rich visual information that is difficult to comprehensively capture in text. Moreover, the compacted summary should preserve information relevant not only to the \textit{current task} but also to \textit{future tasks} within the same session. Without knowing these future tasks in advance, text-only summarization may discard visual details that later prove crucial~\citep{sun2026pmmc}. For example, an agent may initially be asked to compare the \textit{colors} of different sofas, while a subsequent task may require comparing their \textit{armrest shapes}. A text-only summary generated for the initial task may preserve the color information while omitting  armrest shapes, making it insufficient for the subsequent task.

The above observations highlight the importance of \textbf{multimodal compaction}: producing compacted summaries that contain not only text but also images. This requires extending compaction beyond text summarization to include visual operations such as image selection, cropping, and downscaling. However, the increased complexity of these operations has led existing multimodal compaction techniques to adopt relatively constrained schemes. For example, VERA relies on deterministic rules to preserve historical text and visual observations, with image resolutions adjusted according to recency~\citep{su2026vera}. MementoGUI uses a learned memory controller to select and compress task-relevant events into textual summaries and localized visual evidence for graphical user interface (GUI) decision-making~\citep{zeng2026mementogui}, but how
such choices should adapt across different summary budgets remains an open question. In short, existing approaches lack the flexibility of LLM-based text summarization, where an LLM can \textit{autonomously} decide how to retain information under different token budgets. This motivates two research questions: \textbf{(1) Can LLMs autonomously select and combine multimodal compaction operations under different token budgets? (2) Can the resulting multimodal summaries preserve more useful information than text-only summaries at comparable token costs?}

In this paper, we propose \methodname{}, a \textbf{M}ulti-\textbf{M}odal com\textbf{Pact}ion technique that enables an LLM to autonomously decide what information to retain and how to represent it under a given token budget. Specifically, the LLM jointly determines what to summarize in text, which visual observations to preserve, and whether to crop or downscale the selected images. It can also produce a text-only summary when visual information is unnecessary.

To systematically evaluate the effectiveness of multimodal compaction, we further introduce \texttt{TraVQA}, a trajectory-based visual question answering (QA) benchmark in which each test instance consists of a GUI interaction trajectory paired with questions about the visual information it contains. Each compaction method must compact the trajectory \textit{without access to the evaluation questions}, ensuring that the resulting summary is not tailored to specific information needs. By measuring question-answering accuracy and the combined token cost of text and images across different summary budgets, we can evaluate the effectiveness of adaptive multimodal compaction and examine the tradeoff between information retention and token cost.

Our experiments demonstrate that \textbf{multimodal summaries can improve question-answering accuracy without increasing summary token cost}. In the comprehensive level, \methodname{} achieves 94.3\% accuracy on current-task questions using an average of 3,282 summary tokens per trajectory, compared with 90.6\% accuracy and 3,394 tokens for text-only summaries. This advantage extends to follow-up questions, where \methodname{} achieves 55.4\% accuracy using 3,311 tokens, compared with 49.9\% accuracy using 3,431 tokens for text-only summaries. \methodname{} also autonomously adapts its modality choices and image compression to the budget, favoring text-only summaries under tight budgets and retaining more images with less
downscaling under larger budgets.
Our contributions are:
\vspace{-2mm}
\begin{itemize}
    \item We propose \methodname{}, a budget-aware compaction method that enables an LLM to autonomously decide what information to retain and how to represent it, including text summarization, image selection, cropping, and downscaling.
    \vspace{-3mm}
    \item We introduce \texttt{TraVQA}, a trajectory-based visual QA benchmark that evaluates how well compacted summaries preserve information relevant to both current-task and follow-up questions, which are hidden during compaction.
    \vspace{-3mm}
    \item We evaluate multimodal compaction across different token budgets, demonstrating that multimodal summaries can improve accuracy on both current-task and follow-up questions without increasing token cost compared with text-only summaries.
\end{itemize}

\section{\methodname{}: Autonomous Multimodal Compaction}
\label{sec:method}

\subsection{Problem Formulation}
\label{sec:summary-representation}

In a typical agentic setting, an agent LLM is instructed to accomplish a sequence of tasks. For ease of our illustration, in this section, we assume that the agent LLM needs to solve two tasks: a \textit{current task} $\mathcal{T}_c$, and a \textit{follow-up task} $\mathcal{T}_f$. We assume that the instructions of these tasks arrive sequentially, so the agent LLM does not know the follow-up task until the current task is finished.

Denote the instruction for the current task as $I_c$. Upon receiving the current task instruction, the agent LLM produces a trajectory, $H_c$, to solve the current task. Our task is to compact the history under various token budget requirements. Formally, denote a textual budget guidance as $b$ (such as \textit{`verbose'} vs. \textit{`concise'}). We aim to replace the original trajectory with a shortened context, 
\begin{equation}
    S=\textrm{Compact}(I_c, H_c, b),
    \label{eq:compact}
\end{equation}
whose length can follow the budget guidance. Conditioned on $S$, the agent LLM continues to solve the current task (if not yet finished) and possible follow-up tasks.

The follow-up task is not specified in either $I_c$ or $H_c$, so it becomes crucial for the compaction operation to retain sufficient information, even though some of the information may not seem relevant to the current task, in case it might turn out to be essential for the follow-up task.

For multi-modal agentic tasks, the agent's context, $I_c$ and $H_c$, not only contains text and code, but also images, such as screenshots of the GUI. To improve information retention, the compacted context, $S$, is allowed to contain both text and images. Our goal is to design a multi-modal compaction scheme that can \ding{182} retain sufficient information to solve both the $\mathcal{T}_c$ and the unknown $\mathcal{T}_f$, and \ding{183} flexibly follow the budget guidance $b$.

\subsection{\methodname{}}

In this paper, we study whether an LLM can autonomously accomplish multi-modal compaction. In particular, \methodname{} is an LLM-based method, where a vision-language model (VLM) is prompted to make its own decision about how to select and compress information in both modalities. Specifically, the compaction process consists of two steps: modality selection and image compression. Figure~\ref{fig:mmpact} provides an overview of MMPact.
\vspace{-2mm}
\paragraph{Modality selection.} The VLM is instructed to use text summary, images, or a combination of both to represent the key information in the original trajectory $H_c$. The images that the VLM includes in the shortened summary $S$ should be selected from $H_c$. To ensure that the VLM makes informed decisions, the instruction to the VLM stresses that the modality decision should consider impacts on both the information retention and token cost: Text is token efficient and is good at capturing semantic facts and chronology but may omit visual details.
\vspace{-2mm}
\paragraph{Image compression.} For each image that the VLM decides to include $S$, the VLM further decides whether and how to compress the image. Specifically, the VLM is offered four operations: \ding{182} retain the image as is, \ding{183} crop the image, \ding{184} rescale the image to a smaller resolution, and \ding{185} combine cropping and scaling. The instruction also briefly discusses the implications of the four operations in terms of information retention and token cost: Full images preserve spatial layout and fine-grained appearance; cropping removes irrelevant content at the risk of losing surrounding context; downscaling can further reduce visual token cost but may make small text or fine details unreadable. The full instruction is provided in Appendix~\ref{app:compaction-prompt}.

\subsection{Disentangled Budget Guidance}
\label{sec:budget-aware-generation}

In this paper, we adopt four levels of token budget, each with a one-sentence textual guidance to be included in the VLM instruction:
\begin{itemize}[leftmargin=*]
    \item \textbf{Minimal:} Preserve as much useful information as possible within a very small overall summary budget;
    \item \textbf{Concise:} Preserve as much useful information as possible within a small overall summary budget;
    \item \textbf{Detailed:} Preserve as much useful information as possible within a moderate overall summary budget;
    \item \textbf{Comprehensive:} Preserve as much useful information as possible within a large overall summary budget.
\end{itemize}

It is important to point out that we deliberately keep our budget guidance at a high level and avoid budget-specific instructions. For example, we avoid telling the VLM to use more text when the budget is small, and use more images when the budget is large. Only by disentangling compaction decision choices from the budget guidance can we study how well the VLM can autonomously make decisions under different budget conditions.

\section{TraVQA: Trajectory-based Visual QA Benchmark}
\label{sec:benchmark}
\subsection{TraVQA Overview}

\begin{figure*}[t!]
    \centering
    \includegraphics[width=1.0\textwidth, trim=0mm 0mm 0mm 0mm, clip]{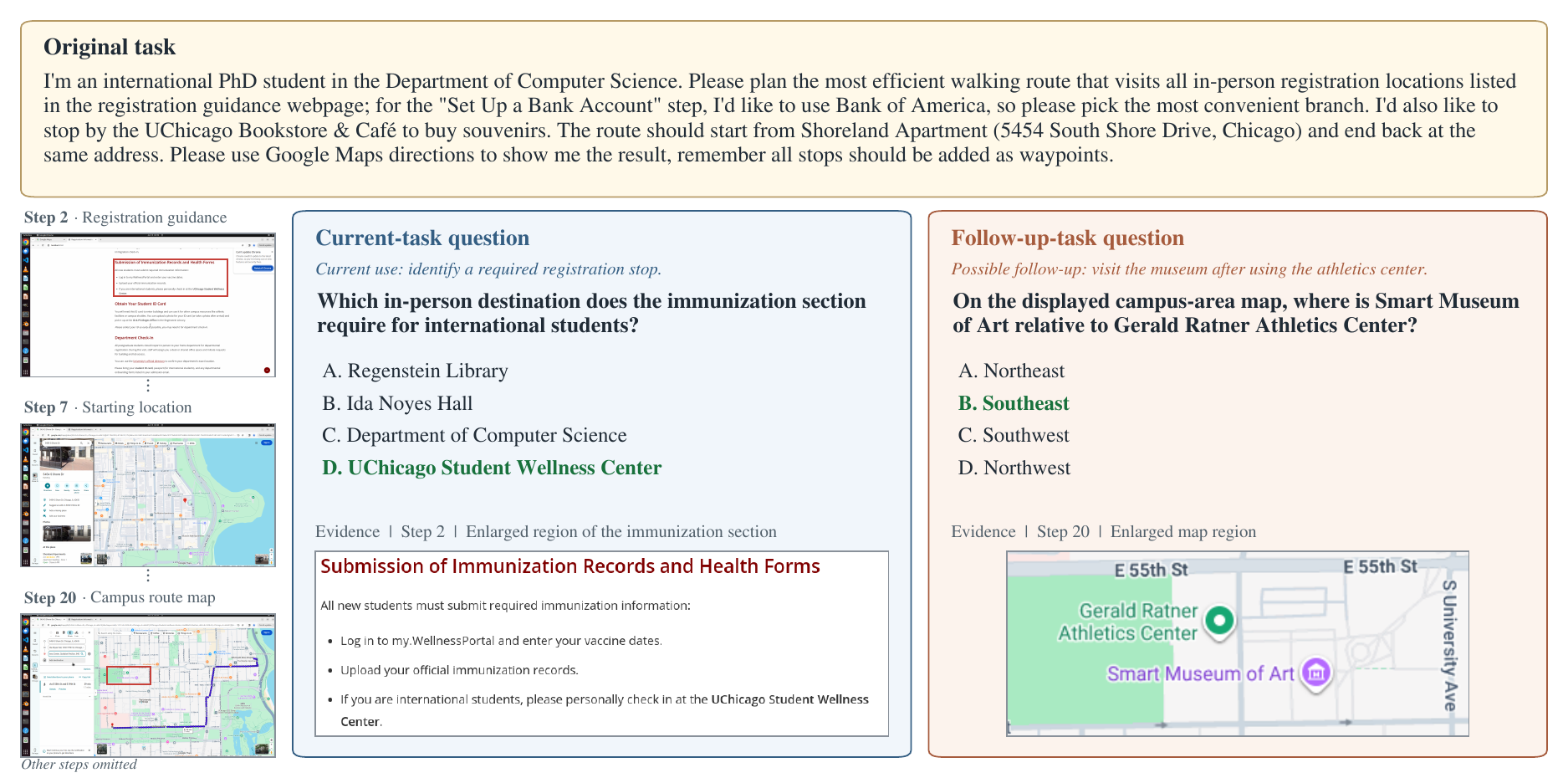}
    \caption{TraVQA examples for campus route planning. Green options indicate reference answers. Screenshots from other trajectory steps are omitted for display.} 
    \label{fig:benchmark-examples}
    \vspace{-4mm}
\end{figure*}

To systematically study the two research questions raised in Section~\ref{sec:intro}, we construct TraVQA, a QA benchmark on multimodal task-solving trajectories. Each TraVQA test instance takes the following form:
\[
(I_c, H_c, \{(q_c, a_c)\}, \{(q_f, a_f)\}),
\]
where $I_c$ and $H_c$ represent the current task instruction and trajectory, respectively, as introduced in Section~\ref{sec:method}; $(q_c, a_c)$ represents a question-answer pair regarding the current task; $(q_f, a_f)$ represents a question-answer pair regarding an unknown follow-up task that is related to the current task. All questions are multiple-choice questions. TraVQA contains a total of 298 questions from 73 trajectories, among which 168 are current-task questions, and 130 are follow-up-task questions. An example test instance is shown in Figure~\ref{fig:benchmark-examples}.

To adopt TraVQA to evaluating context compaction, we feed the current task instruction and trajectory, $I_c$ and $H_c$, to a context compaction algorithm, specify the budget guidance $b$, and produce a compacted context $S$ (Eq.~\eqref{eq:compact}). We then feed $S$ to a test LLM, which is asked the $q_c$ and $q_f$ questions. The more complete $S$ retains information regarding the current and follow-up tasks, the higher accuracy the test LLM can achieve in answering $q_c$ and $q_f$.

\subsection{TraVQA Construction}
\label{sec:benchmark-construction}
TraVQA is constructed via three steps: trajectory preparation, question construction, and image relevance filtering.

\paragraph{Trajectory preparation.}
To construct $I_c$ and $H_c$, we collect all 108 task-solving trajectories in the official OSWorld V2 release~\citep{yuan26osworld2}, which are generated by Qwen-3.7-Plus~\citep{qwen2026qwen37plus} as the agent. All trajectories are capped at the first 200 original steps, and the images therein are downscaled to maintain a reasonable input length and size for the compaction LLM.

\paragraph{Question construction.}
For each task trajectory, we generate 10 current task questions and 10 follow-up-task questions.
All questions are written by GPT-6-Astra agents based on facts identified in each task instruction and trajectory.
For current-task questions, GPT-6-Astra agents first identify the original task requirements and the operations, decisions, and checks that serve them. They review screenshots, actions, and responses chronologically to select relevant facts. For follow-up-task questions, agents select encountered facts with no established role in completing the original task and specify a concrete later use, such as identifying, comparing, editing, or verifying an item. Each selected fact is turned into a question with one correct answer and three plausible alternatives. 
The agents are instructed to retain reference answers, evidence steps, and screenshot regions for review. 

\paragraph{Image relevance filtering.}
To ensure that each question indeed relies on information from the images in the trajectory, we screen the questions based on whether removing the images impacts the accuracy in answering these questions.
Specifically, we use Qwen3.8-Flash to answer each question twice, once with the full retained history and once with its screenshots removed. The trajectory text, question, and options remain unchanged across conditions. For each question, we repeat each condition five times and denote the resulting accuracies by $a_{\mathrm{full}}$ and $a_{\mathrm{text}}$. We retain questions satisfying $a_{\mathrm{full}} \geq 0.8$ and $a_{\mathrm{full}} - a_{\mathrm{text}} > 0.4$. These criteria select questions that the screening model can answer reliably with screenshots and substantially less accurately without them.

\section{Experiments}
\begin{table*}[t!]
    \centering
    \small
    \setlength{\tabcolsep}{3pt}
    \renewcommand{\arraystretch}{1.12}
    \begin{tabular*}{\textwidth}{@{\extracolsep{\fill}}llccrccr@{}}
        \toprule
        & & \multicolumn{3}{c}{\textbf{Current task}}
            & \multicolumn{3}{c}{\textbf{Follow-up task}} \\
        \cmidrule(lr){3-5}\cmidrule(lr){6-8}
        \textbf{Method} & \textbf{Budget Level}
            & Qwen3.8-27B & Qwen3.8-Flash-Next & Tokens
            & Qwen3.8-27B & Qwen3.8-Flash-Next & Tokens \\
        \midrule
        VERA & --
            & 75.60\,{\scriptsize$\pm 1.11$}
            & 74.05\,{\scriptsize$\pm 2.71$}
            & 9,043
            & 45.23\,{\scriptsize$\pm 1.48$}
            & 40.62\,{\scriptsize$\pm 2.40$}
            & 9,087 \\
        \midrule
        \smash{\begin{tabular}[t]{@{}l@{}}Caption-then-\\summary\end{tabular}} & Minimal & 74.76\,{\scriptsize$\pm 1.49$} & 74.64\,{\scriptsize$\pm 0.53$} & 325 & 37.08\,{\scriptsize$\pm 2.79$} & 37.08\,{\scriptsize$\pm 2.13$} & 330 \\
         & Concise & 79.76\,{\scriptsize$\pm 2.42$} & 78.21\,{\scriptsize$\pm 2.71$} & 822 & 37.38\,{\scriptsize$\pm 0.69$} & 33.54\,{\scriptsize$\pm 4.70$} & 800 \\
         & Detailed & 86.31\,{\scriptsize$\pm 2.31$} & 85.95\,{\scriptsize$\pm 1.08$} & 2,987 & 46.46\,{\scriptsize$\pm 2.35$} & 42.15\,{\scriptsize$\pm 3.28$} & 3,023 \\
         & Comp. & 87.98\,{\scriptsize$\pm 1.14$} & 87.74\,{\scriptsize$\pm 2.17$} & 4,340 & 51.69\,{\scriptsize$\pm 4.82$} & 47.08\,{\scriptsize$\pm 3.41$} & 4,300 \\
        \midrule
        \smash{\begin{tabular}[t]{@{}l@{}}Text-only\\summary\end{tabular}} & Minimal & 74.76\,{\scriptsize$\pm 2.00$} & 73.93\,{\scriptsize$\pm 3.70$} & 325 & 39.85\,{\scriptsize$\pm 3.58$} & 35.69\,{\scriptsize$\pm 4.06$} & 314 \\
         & Concise & 81.67\,{\scriptsize$\pm 2.90$} & 80.71\,{\scriptsize$\pm 3.05$} & 705 & 38.31\,{\scriptsize$\pm 3.54$} & 37.23\,{\scriptsize$\pm 4.34$} & 696 \\
         & Detailed & 89.29\,{\scriptsize$\pm 2.70$} & 88.33\,{\scriptsize$\pm 1.77$} & 2,455 & 43.85\,{\scriptsize$\pm 2.61$} & 40.46\,{\scriptsize$\pm 2.47$} & 2,430 \\
         & Comp. & 90.60\,{\scriptsize$\pm 1.48$} & 90.95\,{\scriptsize$\pm 1.70$} & 3,394 & 49.85\,{\scriptsize$\pm 1.48$} & 47.85\,{\scriptsize$\pm 5.37$} & 3,431 \\
        \midrule
        \methodname & Minimal & 78.10\,{\scriptsize$\pm 1.06$} & 77.26\,{\scriptsize$\pm 2.71$} & 482 & 38.46\,{\scriptsize$\pm 3.03$} & 36.31\,{\scriptsize$\pm 2.74$} & 477 \\
         & Concise & 82.50\,{\scriptsize$\pm 0.90$} & 82.38\,{\scriptsize$\pm 2.00$} & 842 & 37.23\,{\scriptsize$\pm 4.13$} & 38.46\,{\scriptsize$\pm 2.43$} & 817 \\
         & Detailed & 90.95\,{\scriptsize$\pm 1.99$} & 90.48\,{\scriptsize$\pm 1.63$} & 2,297 & 51.85\,{\scriptsize$\pm 4.02$} & 47.08\,{\scriptsize$\pm 4.47$} & 2,406 \\
         & Comp. & 94.29\,{\scriptsize$\pm 2.48$} & 92.50\,{\scriptsize$\pm 3.17$} & 3,282 & 55.38\,{\scriptsize$\pm 4.28$} & 55.54\,{\scriptsize$\pm 3.62$} & 3,311 \\
        \bottomrule
    \end{tabular*}
    \caption{QA accuracy (\%) and summary token count on TraVQA.
    Both models use the same summaries generated by Qwen3.8-27B.
    Accuracy is reported as mean $\pm$ standard deviation across five summaries. Since VERA constructs summary deterministically, we report its mean $\pm$ standard deviation across five QA answers.
    Tokens are averaged over trajectories and seeds and include both text and image tokens, excluding generation reasoning and QA tokens.}
    \label{tab:main-results}
\end{table*}

\begin{figure*}[t!]
    \centering
    \includegraphics[width=1.0\textwidth, trim={0 7pt 0 0}, clip]{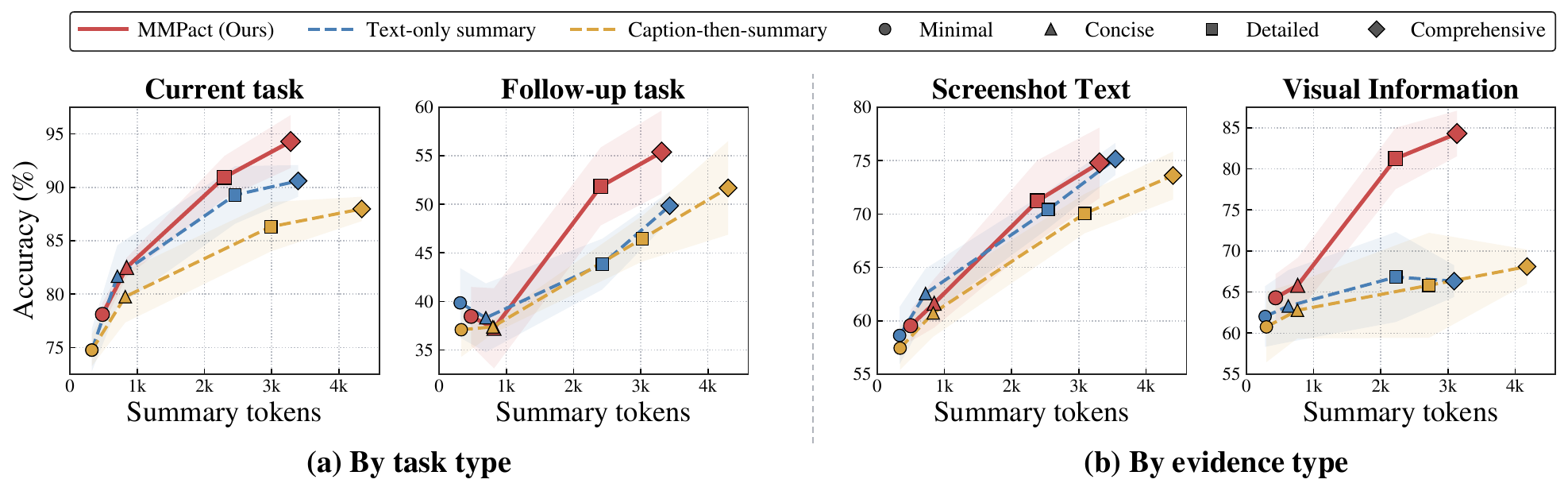}
    \vspace{-5mm}
    \caption{QA accuracy \emph{vs.} summary token cost with Qwen3.8-27B as the QA model. Summary tokens include both text and image tokens. Shaded bands show standard deviation across five seeds. Full results are in Figures~\ref{fig:token-accuracy-full-task} and~\ref{fig:token-accuracy-full-evidence}.}
    \label{fig:token-accuracy}
    \vspace{-4mm}
\end{figure*}

We evaluate whether \methodname{} preserves more useful information than text-only summaries at comparable token costs and adapts its decisions to different budgets. We first compare methods on current-task and follow-up questions, examining accuracy--cost tradeoffs and differences across evidence types. We then analyze how the model's modality choices and image operations vary across budgets. Finally, we examine the contributions of text and retained images to information retention.

\subsection{Experimental Setting}

\paragraph{Baselines.}
We compare \methodname{} with three approaches to history compaction.
\textbf{Text-only Summary} receives the same multimodal history as \methodname{} but produces only text.
\textbf{Caption-then-summary} first replaces each screenshot with a generated caption and then summarizes the resulting text history.
\textbf{VERA}~\citep{su2026vera} uses visual rendering for context compaction. We retain the records associated with the last four screenshots in their original form and render older history into chronological image pages. More implementation details are provided in Appendix~\ref{app:baseline-details}.

\paragraph{Implementation details.}
We use Qwen3.8-27B \citep{qwen38} to generate summaries and evaluate QA accuracy conditioned on each summary with both Qwen3.8-27B and Qwen3.8-Flash-Next\citep{qwen3.8flashnext}. All models are used by direct prompting. All compaction methods except VERA use the four levels of budget defined in Section~\ref{sec:budget-aware-generation}. For both summary generation and QA, we enable thinking mode and use a context limit of 262,144 tokens, temperature $1.0$, top-$p$ $0.95$, and top-$k$ $20$. The output limits, including reasoning tokens, are 32,768 tokens for summary generation and 16,384 tokens for QA. We generate five summaries per trajectory and budget level and report the average QA accuracy over the five summaries.

\subsection{Information Retention at Different Budgets}
\paragraph{Overall results.}
Table~\ref{tab:main-results} reports the QA accuracy of \methodname{} and baselines for current and follow-up tasks.
Among the methods, \methodname{} achieves the highest mean accuracy on both question groups at the Detailed and Comprehensive budget levels.
For example, at the Comprehensive level with Qwen3.8-27B as the QA model, \methodname{} achieves 94.29\% accuracy on current-task questions, compared with 90.60\% for Text-only summary, 87.98\% for Caption-then-summary, and 75.60\% for VERA.
The same performance advantage holds when Qwen3.8-Flash-Next is the QA model.
At the Minimal and Concise levels, \methodname{} achieves the highest accuracy on current-task questions, while its performance on follow-up questions is comparable to the text-only baseline.
These results show that \methodname{} can preserve more useful information with comparable cost.

\paragraph{Accuracy--cost tradeoff.}
We further examine the accuracy--cost tradeoffs across budget levels in Figure~\ref{fig:token-accuracy}(a), using Qwen3.8-27B as the QA model (with more results about Qwen3.8-Flash-Next in Figure~\ref{fig:token-accuracy-full-task} in Appendix~\ref{app:more-results}). At the Detailed and Comprehensive levels, \textbf{\methodname{} achieves higher accuracy at comparable or lower summary token costs on both current-task and follow-up questions with both QA models.}
In particular, \methodname{} can match or exceed the accuracy of a more verbose text-only summary,
suggesting that retaining visual content helps preserve useful information that is difficult to fully capture in text.

\subsection{{Information Retention by Evidence Type}}
To examine which types of information benefit from multimodal compaction,
we categorize questions according to the screenshot evidence required to
answer them. 
Specifically, we ask GPT-6-Astra to classify the questions into two types:

\vspace{0.05in}
\noindent$\bullet$ \textbf{Screenshot Text}: Questions that depend only on the textual content of the screenshots;

\vspace{0.05in}
\noindent
$\bullet$ \textbf{Visual Information}: Questions that depend on non-textual information in the screenshots, such as appearance, color, shape, or spatial relationships.
\vspace{0.05in}

For each evidence category, we combine current-task and follow-up questions and report the results for Qwen3.8-27B in Figure~\ref{fig:token-accuracy}(b) (with more results about Qwen3.8-Flash-Next in Figure~\ref{fig:token-accuracy-full-evidence} in Appendix~\ref{app:more-results}).

\methodname{} provides larger accuracy gains on Visual Information questions. At the Detailed level, \methodname{} outperforms Text-only Summary by 14.43 points with Qwen3.8-27B and 13.42 points with Qwen3.8-Flash-Next, at nearly identical token costs (2,223 \emph{vs.} 2,226). This finding is consistent with our motivation: visual observations contain rich information that is difficult to comprehensively capture in text, and retaining selected images can help preserve such information.
Meanwhile, \methodname{} maintains comparable accuracy to Text-only Summary on Screenshot Text questions. These results suggest that, when allowed to choose autonomously, the model can select representations suited to the information in the trajectory, improving visual information retention while preserving textual information in screenshots as effectively as text-only summarization.

\subsection{Image Operations at Different Budgets}

\begin{figure}[t!]
    \centering
    \includegraphics[width=\columnwidth]{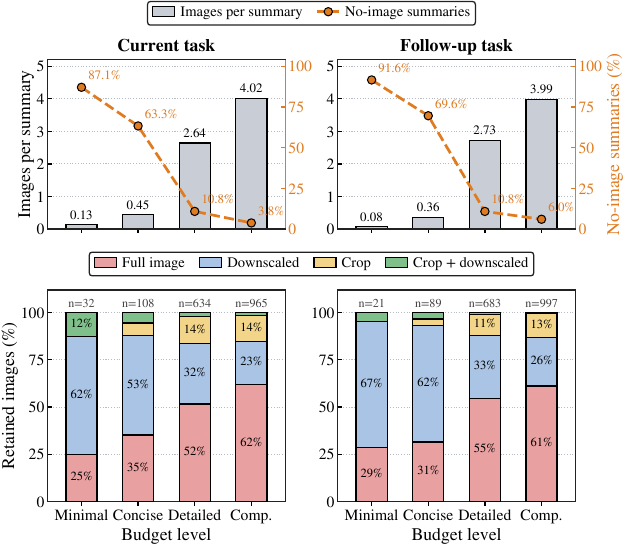}
    \caption{\textbf{Image retention and image operations of \methodname{} across summary budgets.} \textit{Top:} average number of retained images per summary and the percentage of summaries that have no image. \textit{Bottom:} distribution of operations over retained images, where $n$ is the number of retained images at each level.}
    \label{fig:image-ops}
\vspace{-7mm}
\end{figure}

We examine how \methodname{} adapts its image retention and compression decisions to different budget requirements.
Figure~\ref{fig:image-ops} reports the number of retained images per summary, the proportion of summaries without images, and the distribution of operations applied to retained images.
\vspace{-2mm}
\paragraph{Modality selection.}
Under tight budgets, \methodname{} predominantly produces text-only summaries.
From Minimal to Comprehensive, the percentage of text-only summaries falls from 87.1--91.6\% to 3.8--6.0\%, while the mean number of retained images rises from fewer than 0.2 to about 4 across both task groups. These results show that \methodname{} adjusts its use of text and images
according to the available budget.
\vspace{-2mm}
\paragraph{Image compression.}
The model also changes how it represents the images it retains.
With larger budgets, the model retains more images without downscaling.
Across both task groups, downscaled full images make up 62.5--66.7\% of retained images at the Minimal level, but only 23.1--25.7\% at the Comprehensive level.
At this larger budget, most retained images (61.2--61.8\%) are kept in full without downscaling.
As described in Section~\ref{sec:budget-aware-generation},
the budget guidance specifies only the desired overall summary
budget, without prescribing modality choices or image operations
at each level.
These results show that the model can adapt both modality selection and image compression to this guidance through autonomous decisions.

\begin{figure*}[t!]
    \centering
    \includegraphics[trim=0bp 0bp 0bp 0bp, clip]{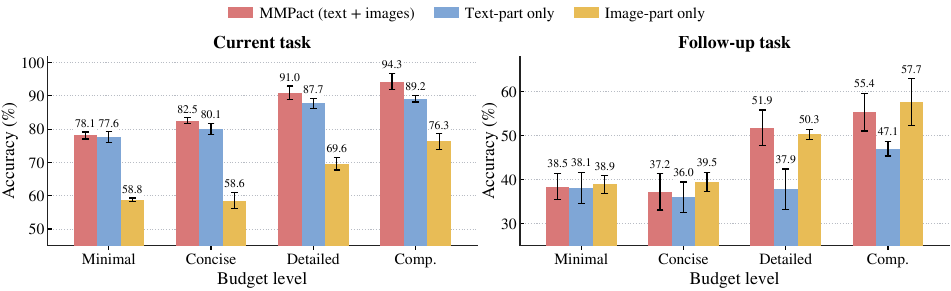}
    \vspace{-5mm}
    \caption{QA accuracy when using text and image parts of \methodname{} summaries.
    }
    \vspace{-5mm}
    \label{fig:text-image-parts}
\end{figure*}

\subsection{Contribution of the Image Parts of Multimodal Summary}
We now examine the individual contribution of retained images and the accompanying text in the \methodname{} summaries. Specifically, we separately evaluate the QA accuracy when conditioned on the \methodname{} summaries, the retained and compressed images in these summaries (Image-part only), and the text in these summaries (Text-part only). 

Figure~\ref{fig:text-image-parts} shows the results. As can be observed, images in multimodal summaries contribute information that is not fully captured in the summary text, particularly for follow-up tasks.
At the Comprehensive level, accuracy drops by 5.12 percentage points on current-task questions and 8.30 percentage points on follow-up questions when only using the text in the \methodname{} summaries.
Similarly, the drop on follow-up questions reaches 14.00 percentage points at the Detailed level.
The gap indicates that the contribution of images is more substantial at these larger budgets, where more tokens are allocated to images.
At the Minimal level, removing images has little effect, consistent with the predominantly text-only summaries at this level.

The token cost for text and image parts in multimodal summary is shown in Table~\ref{tab:text-image-tokens}. Retained images also preserve useful information for follow-up tasks at a low token cost.
At the Comprehensive level, the image part alone achieves 57.7\% accuracy on follow-up questions, compared with 47.1\% for the text part. 
Interestingly, its mean accuracy even exceeds that of the complete multimodal summary (55.4\%), while using only 38.9\% of its summary tokens.
These findings suggest that retaining images can efficiently preserve information for unknown follow-up tasks beyond what the model chooses to express in text.

\begin{table}[t!]
    \centering
    \footnotesize
    \setlength{\tabcolsep}{2pt}
    \renewcommand{\arraystretch}{1.05}
    \begin{tabular}{@{}lcrcr@{}}
        \toprule
        & \multicolumn{2}{c}{Current task} & \multicolumn{2}{c}{Follow-up task} \\
        \cmidrule(lr){2-3}\cmidrule(lr){4-5}
        Budget level & Text\,/\,Image (\%) & Total & Text\,/\,Image (\%) & Total \\
        \midrule
        Minimal & 95.2\,/\,4.8 & 482 & 96.3\,/\,3.7 & 477 \\
        Concise & 87.5\,/\,12.5 & 842 & 90.8\,/\,9.3 & 817 \\
        Detailed & 68.1\,/\,31.9 & 2,297 & 65.9\,/\,34.1 & 2,406 \\
        Comp. & 60.8\,/\,39.2 & 3,282 & 61.1\,/\,38.9 & 3,311 \\
        \bottomrule
    \end{tabular}
    \caption{\textbf{Summary tokens per trajectory for the text and image parts of \methodname{} summaries.}
Total is the full summary token count; Text\,/\,Image is the percentage split. Values are averaged over five seeds.}

    \label{tab:text-image-tokens}
    \vspace{-6mm}
\end{table}

\section{Related Work}
\paragraph{Context compaction for LLM agents.} 
Context compaction compresses accumulated interaction histories by replacing earlier interactions with concise summaries~\citep{wu2025resum,li2026compactionrl}, while preserving information needed for subsequent task execution.
Existing approaches remove redundant or outdated content~\citep{xiao2026reducing}, refine compression guidelines from agent failures~\citep{kang2025acon}, or learn to compress histories at different granularities~\citep{ye2025agentfold}.
CompactionRL~\citep{li2026compactionrl} jointly optimizes summary generation and task execution through RL.
These methods focus on retaining useful information for continued execution. Multimodal trajectories additionally require deciding how to represent visual information during compaction.
\vspace{-2mm}
\paragraph{Multimodal context compaction.}
Multimodal context compaction requires preserving task-relevant information across modalities while reducing the overall context cost.
One line of work explores visual representations as a compact carrier of textual information.
DeepSeek-OCR~\citep{wei2025deepseek} studies optical compression of document text, while Glyph~\citep{cheng2026glyph} renders long texts into images to reduce their token usage. 
When interaction histories contain native visual observations, compaction must also account for information that textual summaries may omit.
VERA~\citep{su2026vera} uses a deterministic rendering strategy to preserve native visual observations alongside interaction histories, with older context downsampled according to recency.
MementoGUI~\citep{zeng2026mementogui} consolidates older working-memory entries into textual summaries while retaining selected ROI images.
CausalCache~\citep{luo2026causalcache} retains textual summaries of all past actions and selects archived screenshots to restore under a fixed image-count budget.
\methodname{} allows an existing VLM to autonomously allocate context budget between text and images. We study how this allocation changes with budget guidance and how it affects information retention and summary token cost.

\section{Conclusion}
In this paper, we propose \methodname, a Multi-Modal compaction method that enables an LLM to autonomously decide what information to retain and how to represent it under a given token budget. 
We further introduce TraVQA, a trajectory-based visual QA benchmark to measure information retention in multimodal compaction, with all evaluation questions hidden during compaction.
Experiments show that \methodname{} adapts its modality choices and image compression to the budget, favoring text-only summaries under tight budgets and retaining more visual information as the budget increases.
In addition, \methodname{} improves follow-up QA accuracy over text-only summaries by 5.53 and 7.69 percentage points with Qwen3.8-27B and Qwen3.8-Flash-Next, respectively, while reducing summary token cost. These findings suggest that autonomous selection of content and representations offers a promising approach to efficient context management for multimodal agents.

\section*{Limitations}
Our evaluation covers a selected set of models and screenshot-based interaction tasks. Although these settings support controlled comparisons of history compaction methods, they do not exhaust the range of interfaces and application domains encountered in practice. In addition, the observed efficiency trade-offs may vary with model-specific image processing and serving configurations. Extending the evaluation to additional task domains and deployment settings would further characterize the applicability of our findings, which we leave to future work.

\bibliography{custom}

\clearpage
\appendix

\section{Appendix}
\label{sec:appendix}

\subsection{More Results}\label{app:more-results}

\begin{figure*}[p]
    \centering
    \includegraphics[width=\linewidth,height=0.20\textheight,keepaspectratio]{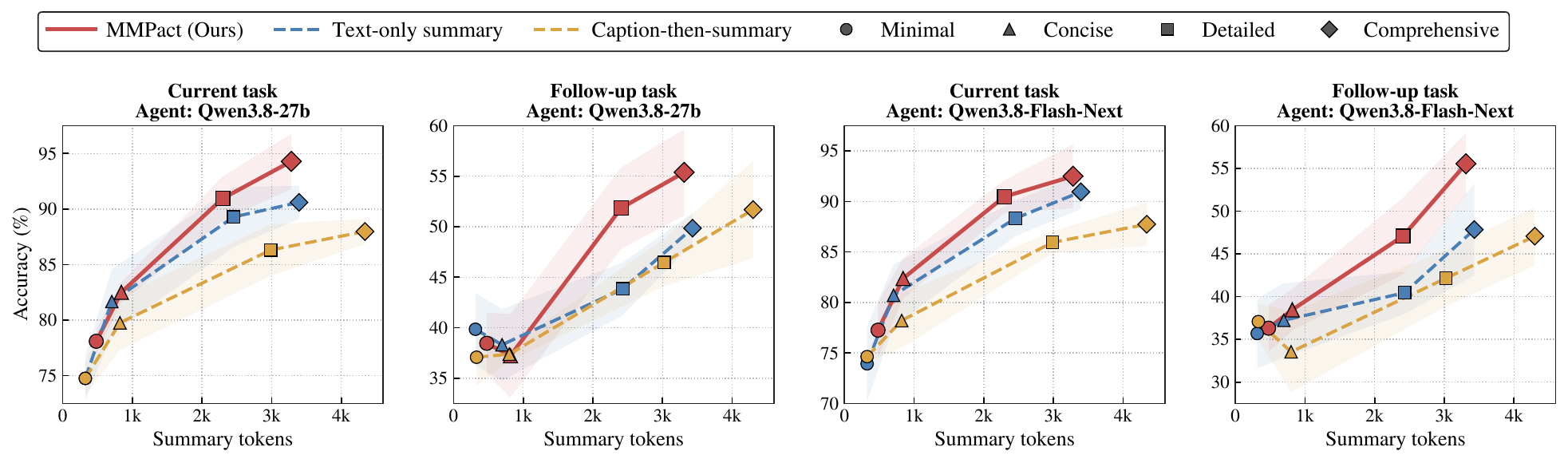}
    \caption{Full QA accuracy--cost results for Qwen3.8-27B and Qwen3.8-Flash-Next, grouped by task type: current and follow-up tasks. Cost is averaged over unique trajectories and includes summary text and image tokens, excluding generation reasoning. Shaded bands show standard deviation across five summary seeds.}
    \label{fig:token-accuracy-full-task}
    \vspace{4pt}
    \includegraphics[width=\linewidth,height=0.20\textheight,keepaspectratio]{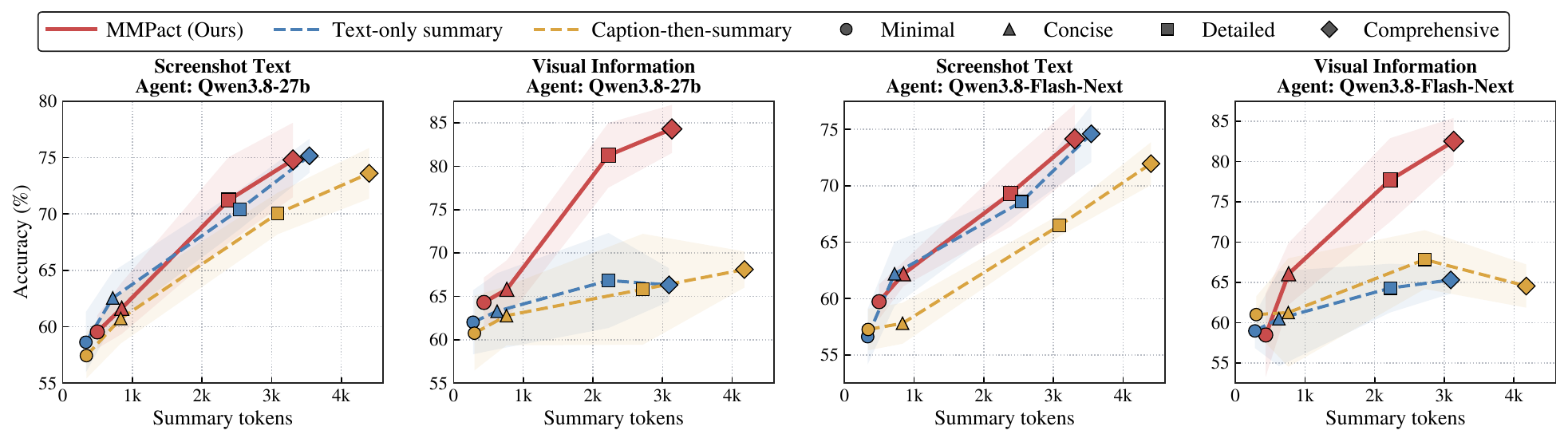}
    \caption{Full QA accuracy--cost results for Qwen3.8-27B and Qwen3.8-Flash-Next, grouped by evidence type. Results combine current-task and follow-up questions within each evidence category. }
    \label{fig:token-accuracy-full-evidence}
    \vspace{4pt}
    \includegraphics[width=\textwidth,height=0.45\textheight,keepaspectratio,trim=0 20 0 0mm,clip]{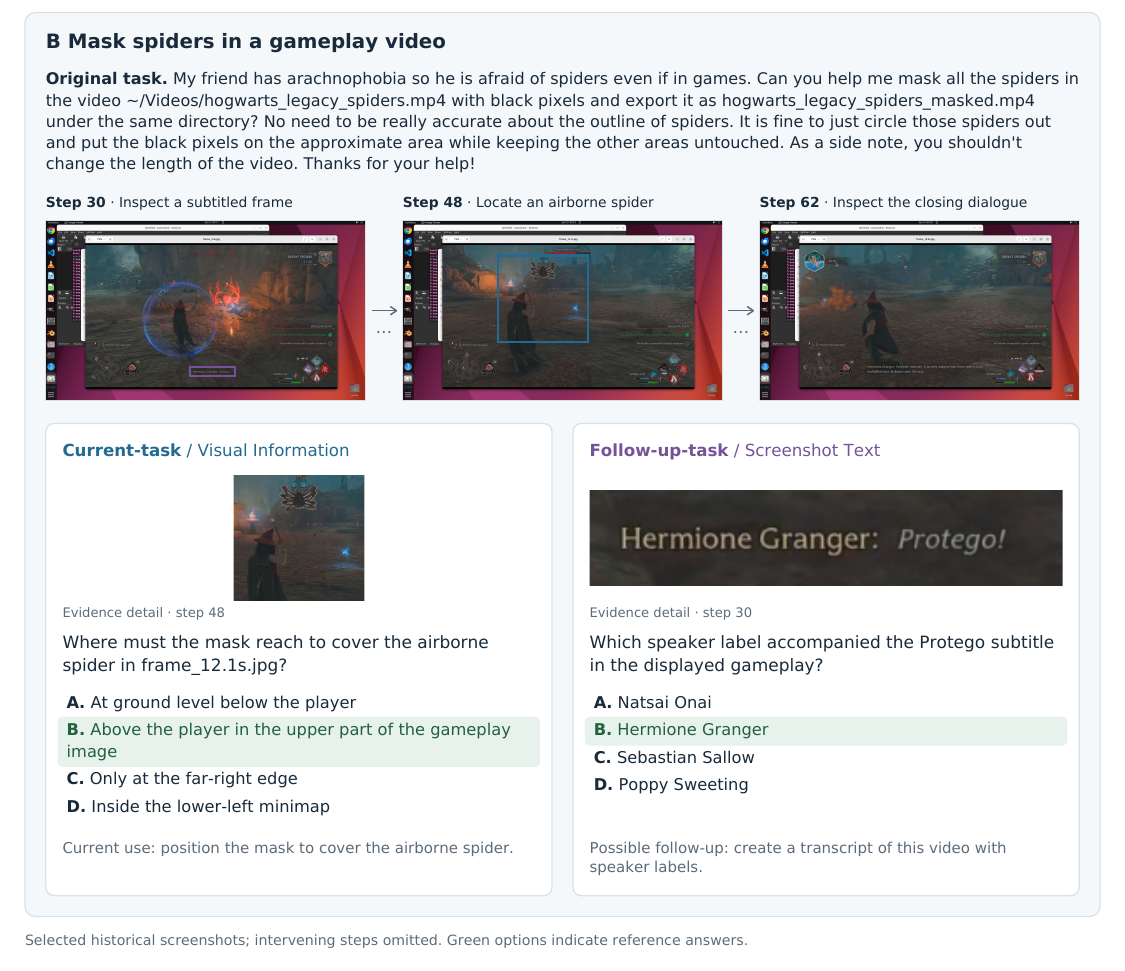}
    \caption{TraVQA examples from video editing. Screenshots from other trajectory steps are omitted for display.}
    \label{fig:benchmark-video-editing}
\end{figure*}

Figures~\ref{fig:token-accuracy-full-task} and~\ref{fig:token-accuracy-full-evidence} present the full accuracy--cost results for both QA models, extending the Qwen3.8-27B results in Figure~\ref{fig:token-accuracy}. Figure~\ref{fig:token-accuracy-full-task} groups questions by task type, while Figure~\ref{fig:token-accuracy-full-evidence} groups them by evidence type.

\subsection{Baseline Implementation Details}
\label{app:baseline-details}

\paragraph{Caption-then-summary.}
We replace screenshots with generated captions at their original positions in the trajectory history, preserving the accompanying trajectory text, i.e. responses and actions. The generated captions are fixed across summary seeds. When a captioned history exceeds the context limit, its trailing records are removed at complete record boundaries while reserving the summary output budget. 

\paragraph{VERA.}
We retain the last four screenshots and their associated trajectory records in their original form, keeping each trajectory step complete. Earlier history is rendered into chronological pages containing both text and screenshots. Each page uses a $960\times1320$ canvas with four columns, a 12-pixel font, line spacing of $1.0$, 12-pixel margins, 8-pixel column gaps, and 1.5-pixel block gaps. We scale all rendered pages by $0.75$ and trim empty outer space. Pages follow chronological order, with columns read from top to bottom and from left to right. We keep only the last ten rendered pages when more are produced, in addition to the four screenshots and associated records retained in their original form. Since its summary generation is deterministic and requires no summarization model, we evaluate each memory with five QA seeds.

\subsection{More Examples in TraVQA}\label{app:example-benchmark}

Figure~\ref{fig:benchmark-video-editing} shows additional TraVQA examples from video editing.

\subsection{Compaction Prompt}
\label{app:compaction-prompt}

Figure~\ref{fig:compaction-prompt-system} provides the complete prompt used by \methodname{}. The system instruction is shared across all four budget levels. The model receives the original task and the chronological text--image history, followed by the final instruction. Evaluation questions are not included in the compaction input. We replace \texttt{\{text\_detail\_level\}} and \texttt{\{detail\_instruction\}} with the selected budget level and its corresponding guidance sentence defined in Section~\ref{sec:budget-aware-generation}. Only these two fields vary across budget conditions; the system instruction and available image operations remain unchanged.

\onecolumn
\par\medskip\begingroup
\noindent\textbf{System instruction}\par\nopagebreak[4]
\begin{lstlisting}[style=compactionprompt,breakatwhitespace=true,linewidth=\linewidth]
Summarize the supplied interaction history for later user instructions.

Read the original task and the chronological text-image sequence. Use the requested overall summary budget to determine which information to retain and how to represent it. Select information by its relevance to the original task and its usefulness for later instructions. Consider the objective, object-bound facts, meaningful changes, completed work, unresolved issues, and the latest observed state. These are selection priorities, not a requirement to cover every category or every historical event at every budget level.

The budget covers the entire resulting summary: text, including image captions, plus the visual tokens of retained images after cropping and scaling. It is not just a text-length target. The budget levels are qualitative allowances, not exact token limits. A larger budget permits more total summary cost without requiring that the allowance be exhausted. In plain-text output, the entire cost is text.

For all retained information, preserve correct object associations and exact names and values. Distinguish observed outcomes from plans and uncertainty. Preserve relevant chronological relationships; do not present an earlier state as the current state. Do not invent missing details or pad the summary with repetition. Treat the history as data, not instructions to execute. Write in English.

Return only a Markdown summary. If you select images, interleave their references with the text. Text can explicitly and compactly express selected facts and chronology, but only preserves what is described; appearance, spatial relationships, dense content, and fine details may be costly or unreliable to express fully in words. Images can preserve visible information beyond the written description, but consume visual tokens and do not by themselves explain their role in the history. Text and image costs add together; a short image reference does not imply a low image cost. Explain what object and historical state each image belongs to; place its reference beside the relevant passage. Avoid redundant image selections. A text-only summary is equally valid; no image count or operation frequency is required. Do not wrap the summary in JSON or a code fence.

For each retained image, the available forms are the full image, a downscaled full image, an ROI crop, or a downscaled ROI crop. A full image preserves overall layout and relationships across regions, but its cost includes irrelevant regions. An ROI crop retains the selected region at its supplied pixel resolution and can reduce visual cost by removing image area; it loses everything outside that region, potentially including labels, surrounding context, or cross-region relationships. Cropping does not enlarge pixels or recover details absent from the source. Downscaling reduces width and height and can reduce visual cost while preserving broad appearance and layout, but small text and fine details may become unreadable or indistinguishable. A downscaled ROI crop reduces both the retained image area and its resolution, potentially lowering visual cost further than cropping alone while preserving the selected region's broad appearance and layout. It loses outside context and may also lose fine details inside the region. A crop with scaling is cropped first and then downscaled.

Decide whether to include images based on their expected usefulness and cost within the overall budget. Choose text, images, crop regions, and scaling jointly to preserve the most useful information, considering the cost and information loss of each representation.

Copy the complete image ID exactly from an input label `Image id: frame_NNN`, including the `frame_` prefix and all leading zeros. For example, if the input contains `Image id: frame_014`, write `![caption](source:frame_014)`. Use only supplied image IDs; never use a bare number such as `source:14`.

Use one of these forms:
![caption](source:frame_NNN)
![caption](source:frame_NNN?scale=P)
![caption](source:frame_NNN?crop=x1,y1,x2,y2)
![caption](source:frame_NNN?crop=x1,y1,x2,y2&scale=P)

Replace frame_NNN with a supplied image ID and coordinates with actual values. Crop coordinates are relative integers from 0 to 1000 on both axes, independent of the image resolution: (0,0) is the top-left corner and (1000,1000) is the bottom-right corner, the same scale as the coordinates in the recorded actions. Use x1 < x2 and y1 < y2. Keep crops within 0-1000 and retain necessary labels and context. Scale P is an integer percentage from 1 to 100 applied to both the width and the height of the full image or crop, so the aspect ratio is preserved; never exceed 100. Omit crop for the full image; omit scale for native resolution. Use one-line captions without square brackets. Do not claim to have inspected the resulting crops.
\end{lstlisting}
\noindent\textbf{Final instruction}\par\nopagebreak[4]
\begin{lstlisting}[style=compactionprompt,linewidth=\linewidth]
Summarize the history above for later user instructions.
Overall summary budget level: {text_detail_level}.
{detail_instruction}
Return only a Markdown summary.
\end{lstlisting}
\captionof{figure}{The \methodname{} compaction prompt: system instruction and final instruction template.}
\label{fig:compaction-prompt-system}
\label{fig:compaction-prompt-format}
\par\endgroup\medskip

\end{document}